\documentclass[letterpaper, 10 pt, conference]{ieeeconf}
\IEEEoverridecommandlockouts
\usepackage[utf8]{inputenc}
\usepackage{amsfonts}
\usepackage{float}
\usepackage[T1]{fontenc}
\usepackage[noadjust]{cite}

\usepackage [autostyle, english = american]{csquotes}
\usepackage[english]{babel}
\MakeOuterQuote{"}
\usepackage{textcomp}
\usepackage{blindtext}
\usepackage{graphicx}
\usepackage{lettrine}
\usepackage{graphicx}
\usepackage{epstopdf}
\usepackage{lipsum}
\usepackage{multicol}
\usepackage{multirow}
\usepackage{textcomp}
\let\labelindent\relax
\usepackage{enumitem}
\usepackage[font=small,skip=4pt]{caption}
\usepackage{amsmath}
\usepackage{color}
\usepackage{mathtools}
\usepackage{booktabs}

\usepackage{array}
\newcolumntype{L}[1]{>{\raggedright\let\newline\\\arraybackslash\hspace{0pt}}m{#1}}
\newcolumntype{C}[1]{>{\centering\let\newline\\\arraybackslash\hspace{0pt}}m{#1}}
\newcolumntype{R}[1]{>{\raggedleft\let\newline\\\arraybackslash\hspace{0pt}}m{#1}}
\usepackage[linesnumbered]{algorithm2e}
\usepackage{bbm}
\usepackage{duckuments}
\usepackage{hyperref}
\hypersetup{colorlinks,linkcolor={black},citecolor={blue},urlcolor={blue}}
\usepackage{balance}
\usepackage[noadjust]{cite}
\usepackage{subcaption}

\usepackage{array}
\newcolumntype{P}[1]{>{\centering\arraybackslash}p{#1}}
\title{Gimballed Rotor Mechanism for Omnidirectional Quadrotors} 

\author{Jann Cristobal, Ahmad Ziad Zain Aldeen, Mohammadreza Izadi, Reza Faieghi
\thanks{The authors are with the Autonomous Vehicles Laboratory, Department of Aerospace Engineering, Toronto Metropolitan University, Toronto, Canada {\tt\footnotesize \{jcristobal, ahmad.zain, mizadi,reza.faieghi\}@torontomu.ca}.}}

\begin{document}

\maketitle

\bstctlcite{IEEEexample:BSTcontrol}

\begin{abstract}
This paper presents the design of a gimballed rotor mechanism as a modular and efficient solution for constructing omnidirectional quadrotors. Unlike conventional quadrotors, which are underactuated, this class of quadrotors achieves full actuation, enabling independent motion in all six degrees of freedom. While existing omnidirectional quadrotor designs often require significant structural modifications, the proposed gimballed rotor system maintains a lightweight and easy-to-integrate design by incorporating servo motors within the rotor platforms, allowing independent tilting of each rotor without major alterations to the central structure of a quadrotor. To accommodate this unconventional design, we develop a new control allocation scheme in PX4 Autopilot and present successful flight tests, validating the effectiveness of the proposed approach.
\end{abstract}


\section{INTRODUCTION}\label{se:intro}

Multirotor uncrewed aerial vehicles (MRUAVs) are utilized in a wide range of applications \cite{mahony2012multirotor,gawel2017aerial}. However, as mission requirements become increasingly complex -- such as surface inspections \cite{steich2016tree,bircher2018receding} and operation in environments with large external disturbances \cite{furieri2017gone} -- conventional MRUAVs face limitations due to their underactuation, which inherently couples position and attitude control.

Omnidirectional MRUAVs have gained significant attention in aerial robotics due to their enhanced maneuverability, improved precision, and greater stability in the presence of disturbances compared to traditional MRUAVs \cite{kamel2018voliro,brescianini2016design,shayan2024nonlinear}. These vehicles can move in any direction without reorienting their body, making them particularly well-suited for challenging environments such as indoor navigation, inspection tasks, and intricate aerial manipulation.

Several designs have been proposed to achieve omnidirectional capabilities, with the two most common approaches being fixed-rotor \cite{brescianini2016design,zhu2024design} and tilt-rotor designs \cite{kamel2018voliro,park2018odar}. Both methods provide a full SE(3) (Special Euclidean Group in 3 dimensions) flight envelope, are overactuated, and require control allocation to distribute forces effectively \cite{falconi2012dynamic}. However, both approaches suffer from inefficiencies due to opposing thrust vectors and increased weight from additional mechanisms, which can impact overall performance.

The works by \cite{brescianini2016design} and \cite{zhu2024design} exemplify fixed-rotor designs that achieve omnidirectional control by strategically orienting thrust vectors, allowing the system to navigate all six degrees of freedom (6-DOF) through individual thrust modulation of each rotor. While fixed-rotor designs are mechanically simple and robust, they suffer from limited versatility, aerodynamic interference, and inefficiencies associated with bidirectional propellers \cite{maier2018bidirectional}.

The works by \cite{kamel2018voliro} and \cite{park2018odar} are examples of tilt-rotor designs that achieve omnidirectional control by dynamically adjusting thrust vectors through variations in both tilt angles and rotor velocities, enabling full 6-DOF maneuverability. While tilt-rotor designs offer high maneuverability, precise manipulation, and enhanced stability, further improvements can be made by addressing challenges related to mechanical complexity, modularity, and responsiveness.

Existing tilt-rotor designs are mechanically complex and heavy \cite{kamel2018voliro,park2018odar,wu2022design,zheng2020tiltdrone, myeong2018development}. In \cite{zhu2024design} and \cite{zheng2020tiltdrone}, large modifications to the central body are done to accommodate the additional parts necessary to achieve full actuation. The use of DC motors paired with an encoder in \cite{kamel2018voliro} adds to the overall mass. It also suffers from slow response time and limited torque. The use of gears and lever arms in \cite{myeong2018development} introduces an additional point of failure and increases complexity and weight.

The design presented in this paper addresses the above limitations.
By utilizing gimballed rotors, the system remains simplified, modular, and efficient.
Built on conventional quadrotor frame, this design requires the addition of only four servo motors that are directly connected to the rotor bases. Its high modularity allows for easy repairs and modifications by simply altering the rotor platforms to accommodate additional actuators. Additionally, the use of small, fast, high-torque servo motors enhances system responsiveness while reducing energy consumption.

Another key aspect of our design is its focus on accessibility, utilizing off-the-shelf components and readily available manufacturing methods. We have deliberately chosen widely used platforms such as PX4 Autopilot (PX4) and Robot Operating System (ROS) to ensure that developers can easily replicate and modify the design.
Leveraging these established technologies facilitates the accessibility of  omnidirectional quadrotors and accelerating the advancement and adoption of these aerial platforms.

\section{MECHANISM DESIGN}

\subsection{Design Requirements}
The objective of the gimballed rotor mechanism is to merge the advantages of tilt-rotor and fixed-rotor designs, enhancing the maneuverability of conventional quadrotors while maintaining structural simplicity, modularity, and robustness. 
By incorporating the gimballed mechanism on all four arms of a conventional quadrotor, the vehicle rotors will have tilting capabilities along longitudinal and lateral axes.
As such, the vehicle can achieve omnidirectional control without the need to modify the vehicle’s central frame.

The final design must satisfy the following component-level requirements (CLRs). Note that the following functional specifications are tailored for an off-the-shelf quadrotor (Holybro X500), which we intend to modify into an omnidirectional quadrotor.

\begin{itemize}
\item CLR-1: The gimbal assembly must withstand a maximum thrust of $13N$.  
\item CLR-2: The gimbal mechanism must achieve a tilting angle range of $\pm 20^{\circ}$. This was chosen based on the requirement of $1g$ acceleration laterally and the vehicle's thrust-to-weight ratio of $3:1$. 
\item CLR-3: The gimbal mechanism must generate a minimum torque of $0.3Nm$ based on an average disturbance force of $5N$ and a moment arm length of $6cm$ from the rotor tip to the axis of rotation.  
\item CLR-4: The gimbal mechanism must have a tilting speed greater than $7.85 \frac{rad}{s}$. This value is selected from \cite{kamel2018voliro}.  

\end{itemize}

\begin{figure}[t]
    \centering
        \includegraphics[trim={5cm, 15cm, 2cm, 10cm}, clip, width = \linewidth]{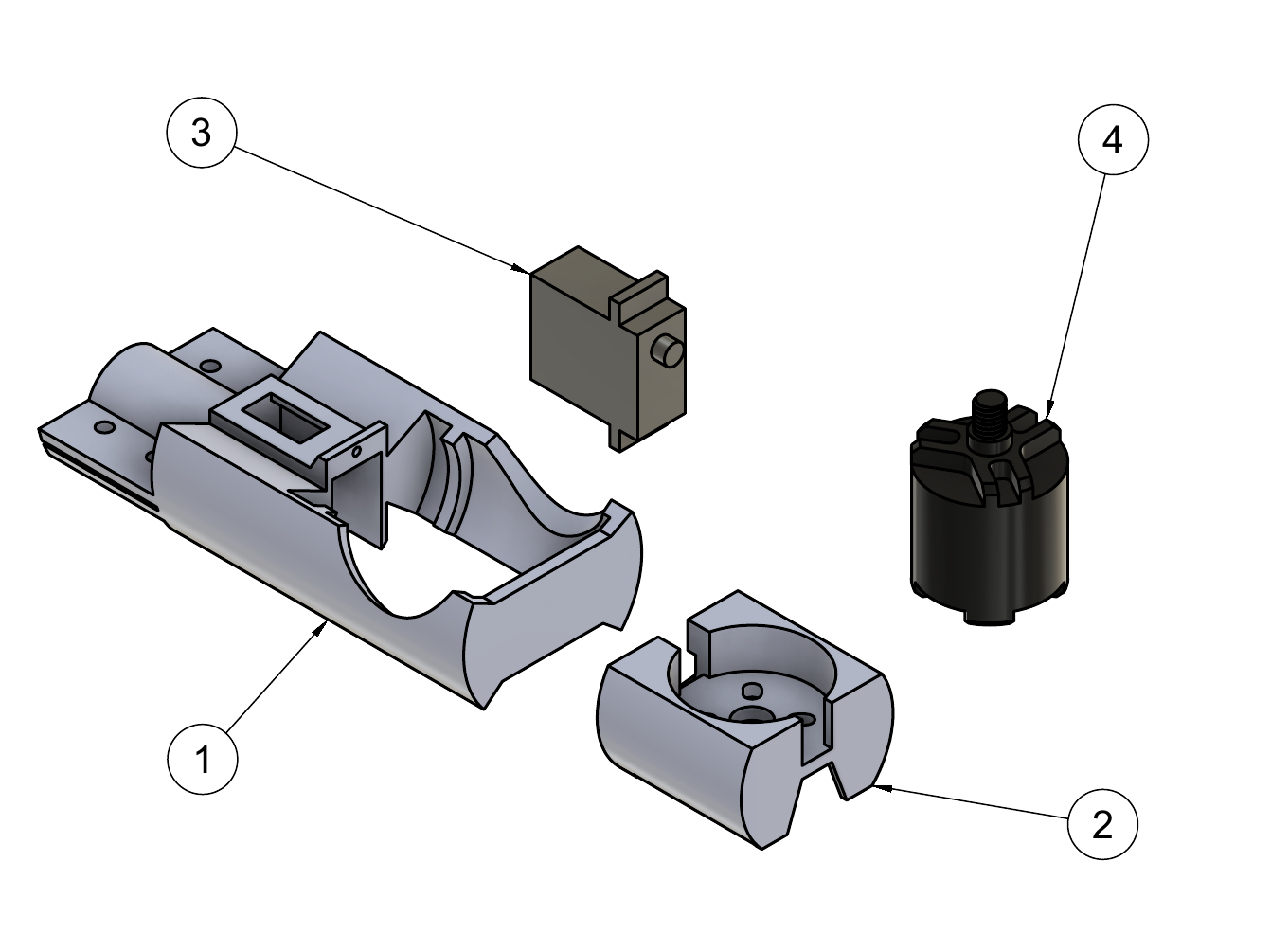}
    \caption{Gimballed Rotor Mechanism Exploded View}
    \label{fig:gimballedrotor}
\end{figure}

\subsection{Design Description}
Figure \ref{fig:gimballedrotor} shows the proposed gimballed rotor mechanism to meet the above design requirements. There are two key components to this assembly. First is the base clamp (Fig \ref{fig:gimballedrotor}.1), where the servo motor (Fig \ref{fig:gimballedrotor}.3) is housed and grooves are located to allow for the rotation to occur smoothly. Second is the rotating platform (Fig \ref{fig:gimballedrotor}.2), where the rotor (Fig \ref{fig:gimballedrotor}.4) is placed on top. 

One challenge with tilt-rotor designs is the presence of large oscillations when the rotors operate at high velocities and the servo motor attempts to maintain a desired angle. The proposed configuration of the clamp, rails, and tracks, as shown in Fig. \ref{fig:gimballedrotor}, proved effective for mitigating these oscillations, provided that their fit is appropriately tuned considering factors such as the thrust levels, servo response, and structural stiffness, damping, and friction.

The width of the gimballed rotor mechanism measures $45 mm$, a modest increase from the $35 mm$ of the original rotor platform included in the original quadrotor. This ensures minimal aerodynamic interference, preventing significant thrust loss. At the same time, the design achieves a $\pm40^\circ$ range of motion and a lateral acceleration of $10 m/s^2$, exceeding the requirements outlined in CLR-2.

The mechanism is straightforward to assemble and disassemble using common tools such as a screwdriver and pliers. The assembly steps include (1) attaching the base clamp to the quadrotor arm, replacing the original rotor platform, (2) inserting the servo motor into the base clamp and securing it with the fastener, (3) inserting the rotating platform into the base clamp, ensuring it securely attaches to the servo arm, and (4) installing the rotors using the included fasteners.

Unlike existing designs that require extensive modifications to the central body -- rendering the entire system inoperable in the event of a crash or failure -- this approach enhances accessibility and repairability for all components. 

Another advantage of the proposed design is that the servo motor is directly connected to the rotor base, eliminating the need for complex gear systems, as is the case with most existing omnidirectional MRUAVs. This direct-drive approach reduces weight and enhances robustness by minimizing potential points of failure.

\subsection{Prototype Fabrication}
For structural fabrication, we chose 3D printing, using carbon-fiber-reinforced high-temperature polyamide as the primary material due to its high strength, low density, durability, and heat resistance.
The specific carbon-fiber-reinforced high-temperature polyamide we used is rated to withstand normal stresses up to $125 \text{ MPa}$ under bending loads, ensuring structural integrity during operation. Compared to common materials such as polylactic acid, it maintains a low density of approximately $1.06 \text{ g/cm}^3$, preserving a high thrust-to-weight ratio essential for maximizing flight time. Additionally, its robustness against external disturbances and high heat deflection temperature of approximately $194^{\circ}C$ make it well-suited for handling the thermal loads generated by the servo motors and rotors.

As for the servo motor, we chose the SAVOX SV-1232MG Micro Servo for its durable and long-lasting metal gear train. Its all-metal case enhances heat dissipation, enabling high-voltage operation with fast response times. At its maximum operating voltage of $7.4V$, it delivers a torque of $5.0 \text{ kg-cm}$ and a speed of $20.94 \frac{\text{rad}}{s}$. Weighing only $23g$ and measuring $23 \text{mm} \times 12 \text{mm} \times 27.3 \text{mm}$, this servo provides the necessary actuation power, while keeping the gimbal lightweight and compact.

\subsection{Design Comparisons} 
Table \ref{tb: design-comparison} compares our proposed gimballed rotor mechanism with existing approaches for omnidirectional MRUAV design.
As discussed in Section \ref{se:intro}, existing designs typically require extensive modifications to the vehicle’s central frame, increasing design complexity and weight.

Since the range of motion of the mechanism is $\pm 40^\circ$, our design does not achieve the full SE(3) flight capability of other approaches, which rely on a large number of actuators and substantial frame modifications to enable unrestricted motion. However, our approach provides a significantly reduced actuation requirement (4 motors + 4 servos) and minimal frame modifications (rated 1/5 in complexity).

\begin{table*}[t]
\begin{minipage}{\textwidth}
\centering
\caption{Existing Omnidirectional Multirotors Design Comparison}
\label{tb: design-comparison}
\begin{tabular}{l c c c c}
  \toprule
  Design Approach & Dynamic Thrust Vector & Flight Envelope & Frame Modification\footnote{The higher rating means that the design complexity is higher relative to the conventional multirotor design.} & Number of actuators\\
  \midrule
  Voliro \cite{kamel2018voliro} & YES & Entire SE(3) & 2/5 & 12 motors\\
  ODAR \cite{park2018odar} & YES & Entire SE(3) & 4/5 & 16 motors\\
  Omnicopter \cite{brescianini2016design} & NO & Entire SE(3) & 3/5 & 8 motors\\
  FAM \cite{zhu2024design} & NO & Entire SE(3) & 5/5 & 12 motors\\
  Gimballed Rotor & YES & Partial SE(3) & 1/5 & 4 motors + 4 servos\\ 
  \bottomrule
\end{tabular}
\end{minipage}
\end{table*}

\section{OMNIDIRECTIONAL QUADROTOR MODEL}\label{se:modeling}
\begin{figure}[t]
    \centering
        \includegraphics[trim={15cm, 0cm, 15cm, 0cm}, clip, width = \linewidth]{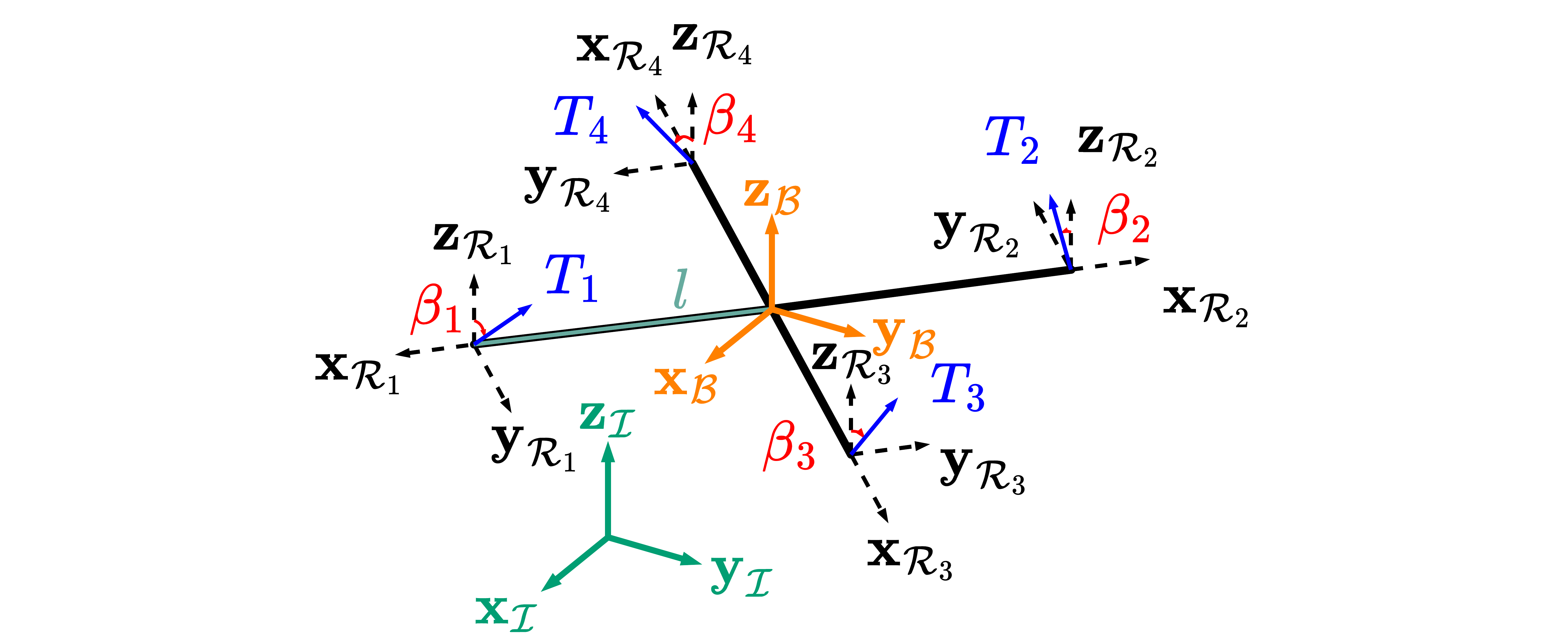}
    \caption{Coordinates frames and parameters setup in our modeling approach for omnidirectional quadrotor with gimballed rotors}
    \label{fig:modeling}
\end{figure}
This section presents the flight dynamics modeling of an omnidirectional quadrotor built using gimballed rotor mechanisms.

\subsection{Model Assumptions}
We assume that the vehicle follows an X-configuration, has a rigid and symmetric structure, and that the axis of rotation of the tilting mechanism aligns with the center of mass. Additionally, we consider the servo dynamics to be sufficiently fast to be approximated as instantaneous. Lastly, we assume that the thrust forces of each rotor act independently, without mutual interference. However, it is important to note that this assumption holds only for small tilting angles. As the tilt angle increases, thrust-induced lateral forces become more significant, leading to increased aerodynamic interference.

\subsection{Rigid Body Dynamics}

Let $\mathcal{I}=\left\{{\bf{x}}_\mathcal{I},{\bf{y}}_\mathcal{I},{\bf{z}}_\mathcal{I} \right\}$ denote the inertial coordinates frame, $\mathcal{B}=\left\{{\bf{x}}_\mathcal{B},{\bf{y}}_\mathcal{B},{\bf{z}}_\mathcal{B} \right\}$ the body frame, as shown in Fig. \ref{fig:modeling}.
Let ${\boldsymbol{\xi}}=\left[x,y,z\right]^T$ denote the vehicle position, and $\boldsymbol{\eta}  = {[\phi,\theta,\varphi ]^T}$ denote the attitude, where $ - \pi  < \phi  \le \pi $, $ - \frac{\pi }{2} \le \theta  \le \frac{\pi }{2}$, and $ - \pi  < \psi  \le \pi $ are the Euler angles representing roll, pitch, and yaw in the yaw-pitch-roll sequence. Let $\boldsymbol{\upsilon}=\left[u,v,w\right]^T$ denote the linear velocity, ${\boldsymbol{\omega}}=\left[p,q,r\right]^T$ the angular velocity, and ${_\mathcal{I}}{\bf{R}}_{\mathcal{B}}$ the rotation matrix from $\mathcal{B}$ to $\mathcal{I}$.

The translational dynamics can be expressed as follows
\begin{equation}
\label{eq:translationalDyn}
\ddot{\boldsymbol{\xi}} = {\bf{g}} + \frac{1}{m}\left({_\mathcal{I}}{\bf{R}}_{\mathcal{B}}{\bf{f}}_P^\mathcal{B}-{\bf{A}}_T{\boldsymbol{\upsilon}}+{\bf{f}}_D\right),
\end{equation}
where ${\bf{g}}=[0,0,9.81]^T\;m/s^2$ is the gravity vector, $m$ is the vehicle mass, ${\bf{A}}_T$ is the translational drag coefficient, and ${\bf{f}}_P^\mathcal{B}$ and ${\bf{f}}_D$ represent propulsive and disturbance forces.
The superscript $\mathcal{B}$ is to highlight ${\bf{f}}_P$ is expressed in the body frame of reference located at the vehicle's center of mass. 
This will become useful in the formulation of the propulsive forces and moments to be detailed shortly.

The rotational dynamics of the vehicle can be expressed as follows
\begin{equation}
\label{eq:rotationalDyn}
\begin{array}{c}
     \dot{\boldsymbol{\eta}}={\bf{H}}\left(\boldsymbol{\eta}\right)\boldsymbol{\omega},  \\
     {\boldsymbol{\dot \omega }} = {{\bf{J}}^{-1}}\left(-\boldsymbol{\omega} \times {\bf{J}}{{\boldsymbol{\omega }} + {{\boldsymbol{\tau}}_P^{\cal B}} - \bf{A}}_R\boldsymbol{\omega} +{\boldsymbol{\tau}}_d\right),
\end{array}
\end{equation}
where $ \times $ indicates the cross product, ${\bf{J}}$ is the vehicle inertia matrix, ${\bf{A}}_R$ is the rotational drag coefficient, ${\boldsymbol{\tau}}_P^\mathcal{B}$ and ${\boldsymbol{\tau}}_D$ represent propulsive and disturbance moments, and
\begin{equation}
\label{eq:H}
{\bf{H}}\left( {\boldsymbol{\eta }} \right) = \left[ {\begin{array}{*{20}{c}}
1&{\sin \phi \tan \theta }&{\cos \phi \tan \theta }\\
0&{\cos \phi }&{ - \sin \phi }\\
0&{{{\sin \phi } \mathord{\left/
 {\vphantom {{\sin \phi } {\cos \theta }}} \right.
 \kern-\nulldelimiterspace} {\cos \theta }}}&{{{\cos \phi } \mathord{\left/
 {\vphantom {{\cos \phi } {\cos \theta }}} \right.
 \kern-\nulldelimiterspace} {\cos \theta }}}
\end{array}} \right].
\end{equation}

\subsection{Forces and Torques}
To formulate the propulsive forces and moments ${\bf{f}}_P$ and ${\boldsymbol{\tau}}_P$, we start by defining the $i$-th rotor coordinates frames $\mathcal{R}_i$ with the following orientation with respect to $\mathcal{B}$, also illustrated in Fig. \ref{fig:modeling},

\begin{equation}
    \label{eq:rotmat-z}
    {\bf{R}_{z}}\left( {\alpha} \right) = \left[ {\begin{array}{*{20}{c}}
\cos \alpha &-\sin \alpha&0\\
\sin \alpha&\cos \alpha&0\\
0&0&1
\end{array}} \right].
\end{equation}

\begin{equation}
\label{eq:rotorFramesSetup}
\begin{array}{l}
{_\mathcal{B}}{\bf{R}}_{\mathcal{R}_1} ={\bf{R}_{z}}\left( {315^\circ} \right),\\
{_\mathcal{B}}{\bf{R}}_{\mathcal{R}_2} = {\bf{R}_{z}}\left( {135^\circ} \right),\\
{_\mathcal{B}}{\bf{R}}_{\mathcal{R}_3} = {\bf{R}_{z}}\left( {45^\circ} \right),\\
{_\mathcal{B}}{\bf{R}}_{\mathcal{R}_4} = {\bf{R}_{z}}\left( {225^\circ} \right).
\end{array}
\end{equation}

Next, we represent the thrust, torque, and tilting angles about the longitudinal axis of the $i$-th rotor by $T_i$, $Q_i$, and $\beta_i$. 
We note that
\begin{equation}
\label{eq:thrust}
    T_i \approx k_T \Omega_i^2, \; Q_i \approx k_Q \Omega_i^2,
\end{equation}
where $k_T$ and $k_Q$ are rotor coefficients, and $\Omega_i$ is the angular velocity of rotor with opposite signs $\Omega_{1,2} > 0$ and $\Omega_{3,4} < 0$. 
The propulsive force due to $T_i$, denoted by ${\bf{f}}_{P_i}$ is expressed in $\mathcal{R}_i$ as follows

\begin{equation}
\label{eq:rotorForces}
\begin{array}{ll}
  {\bf{f}}_{P_i}^{\mathcal{R}_i} = \left[f_{{lon}}^{\mathcal{R}_i}, f_{{lat}}^{\mathcal{R}_i},f_{{ver}}^{\mathcal{R}_i} \right]^T,  \\
  {\bf{f}}_{P_i}^{\mathcal{R}_i} = T_i 
  \begin{bmatrix}
  0\\
  \sin\left({\beta_i}\right)\\
  \cos\left(\beta_i\right)    
  \end{bmatrix}
  ,   & i = 1,2,3,4 \\
\end{array}
\end{equation}
Therefore, the individual, ${\bf{f}}_{P_i}^{\mathcal{B}_{i}}$, and total, 
 ${\bf{f}}_P^{\mathcal{B}}$, propulsive force expressed in $\mathcal{B}$ is
\begin{equation}
\label{eq:totalPropulsiveForces}
\begin{array}{ll}
{\bf{f}}_{P_i}^{\mathcal{B}_{i}} = {_{\mathcal{B}}}{\bf{R}}_{\mathcal{R}_i}{\bf{f}}_{P_i}^{\mathcal{R}_i}, & i = 1,2,3,4\\
{\bf{f}}_P^{\mathcal{B}} = \sum_{i=1}^4 {\bf{f}}_{P_i}^{\mathcal{B}_i}.
\end{array}
\end{equation}

To construct $\boldsymbol{\tau}_P^\mathcal{B}$, let $l$ represent the distance of rotor from the vehicle centre of mass. 
Then, the moments due to the rotors' thrusts and torques take the following form 
\begin{equation}
\label{eq:propulsiveMoments}
\begin{array}{lll}
    \tau_{P_x}^{\mathcal{B}} & = & {\frac{\sqrt{2}}{2}l}\left( -f_{{ver}}^{\mathcal{B}_{1}} + f_{{ver}}^{\mathcal{B}_{2}} + f_{{ver}}^{\mathcal{B}_{3}} - f_{{ver}}^{\mathcal{B}_{4}}\right), \\
    \tau_{P_y}^{\mathcal{B}} & = & {\frac{\sqrt{2}}{2}l}\left( -f_{{ver}}^{\mathcal{B}_{1}} + f_{{ver}}^{\mathcal{B}_{2}} - f_{{ver}}^{\mathcal{B}_{3}} + f_{{ver}}^{\mathcal{B}_{4}}\right), \\
    \tau_{P_z}^{\mathcal{B}} & = & {k_Q}\left( -f_{{ver}}^{\mathcal{B}_{1}} - f_{{ver}}^{\mathcal{B}_{2}} + f_{{ver}}^{\mathcal{B}_{3}} + f_{{ver}}^{\mathcal{B}_{4}}\right) \\ &   & + {l}\left( -f_{{lat}}^{\mathcal{B}_{1}} + f_{{lat}}^{\mathcal{B}_{2}} + f_{{lat}}^{\mathcal{B}_{3}} - f_{{lat}}^{\mathcal{B}_{4}}\right).
\end{array}
\end{equation}

It is noteworthy that our systematic approach in constructing the propulsive forces and moments can be easily extended to other vehicle configurations with an arbitrary number of rotors with gimballed rotor mechanisms.

\section{CONTROL}\label{se:control}
\begin{figure*}[t]
    \centering
        \includegraphics[trim={10cm, 0cm, 0cm, 0cm}, clip, width = 0.8\linewidth]{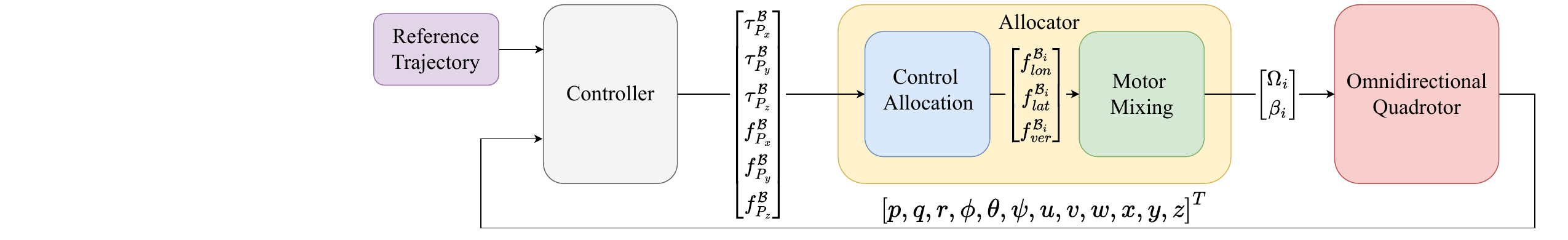}
    \caption{Overview of the control system framework for the omnidirectional quadrotor.}
    \label{fig:blockdiagram}
\end{figure*}

Common autopilot programs, such as PX4 \cite{px4}, which is widely adopted by the research community, do not natively support omnidirectional quadrotors. Therefore, to control the omnidirectional quadrotor equipped with gimballed rotor mechanisms, we modified PX4 to incorporate a customized control strategy.

\subsection{Position and Attitude Control}
The standard PX4 control architecture for multicopters involves a cascaded control structure, where the outer loop governs position control, generating thrust and attitude setpoints, while the inner loop controls the attitude to achieve these setpoints while passing the thrust setpoint as-is.

To decouple position and attitude controllers for omnidirectional quadrotors, we modified the PX4 position control module, \texttt{mc\_pos\_control}, to continuously generate a fixed attitude setpoint of $\boldsymbol{\eta} = {[0,0,0]^T}$ for maneuvers that require translational motion at level attitude.

The standard position controller in PX4 takes trajectory setpoints and generates a 3D thrust vector. This vector is then used as the $\mathbf{z}_\mathcal{B}$ axis of the body frame, generating an attitude setpoint using the \texttt{bodyzToAttitude} function. The total thrust is then calculated as the magnitude of the thrust vector, resulting in a setpoint only on the $\mathbf{z}_\mathcal{B}$ axis.

This implementation allowed for easy separation of attitude and position control. We overrode the \texttt{bodyzToAttitude} function, publishing the internal 3D thrust vector as-is, and setting the attitude setpoint to the quaternion representation of a level attitude.

The PX4 position and attitude control loops rely on proportional-integral-derivative (PID) controllers. For our experiments, we did not modify the tuning of these parameters to ensure that the observed behavior was representative of standard PX4 performance with minimum change.

\subsection{Control Allocation}
One key difference in controlling omnidirectional quadrotors compared to conventional ones lies in control allocation because they have more actuators and greater control authority, requiring a more complex allocation strategy.
The control allocation in PX4 is handled by the \texttt{control\_allocator} module, which takes force and torque setpoints, each as a 3D vector as an input and generates motor and servo commands as output. In this module, we implemented the new control allocation, which will be detailed shortly, by creating two new classes inherited from the following classes:
\begin{itemize}
    \item \texttt{ActuatorEffectiveness} which constructs the control allocation matrix using rotors positions, coefficients of thrust and torque, and neutral thrust axis.
    \item \texttt{ControlAllocation} which takes the effectiveness matrix as an input, calculates its pseudo-inverse, and normalizes the mixing matrix as PX4 requires actuator outputs to be in the range $[-1, 1]$. This class also allocates the controls for the servos and motors, adding trims and ensuring the servo and mechanism limits are not exceeded.
\end{itemize}

To formulate our control allocation approach, let us define
\begin{equation}
        \textbf{v} = \left[\tau_{P_x}^{\mathcal{B}}\quad\tau_{P_y}^{\mathcal{B}}\quad\tau_{P_z}^{\mathcal{B}}\quad f_{P_x}^{\mathcal{B}}\quad f_{P_y}^{\mathcal{B}}\quad f_{P_z}^{\mathcal{B}}\right]^{T},
\end{equation}
\begin{equation}\label{eq:u}
            \textbf{u} = \left[ f_{{lon}}^{\mathcal{B}_{1}} \quad f_{{lat}}^{\mathcal{B}_{1}}\quad f_{{ver}}^{\mathcal{B}_{1}},\quad ... \quad, f_{{lon}}^{\mathcal{B}_{4}} \quad f_{{lat}}^{\mathcal{B}_{4}}\quad f_{{ver}}^{\mathcal{B}_{4}} \right]^{T}.
\end{equation}
It follows from Section \ref{se:modeling} that 
\begin{equation}
    \textbf{v} = \textbf{B}\textbf{u}
\end{equation}
where $\textbf{B}$ is the control effectiveness matrix given by
\begin{equation}
        \textbf{B} = \begin{bmatrix}
        0&0&0&0&0&0\\
        0&0&-l&k_{x}&k_{x}&0\\
        -k_{x}l&-k_{x}l&-k_{Q}&0&0&1\\
        0&0&0&0&0&0\\
        0&0&-l&-k_{x}&-k_{x}&0\\
        k_{x}l&k_{x}l&-k_{Q}&0&0&1\\
        0&0&0&0&0&0\\
        0&0&-l&-k_{x}&k_{x}&0\\
        k_{x}l&-k_{x}l&k_{Q}&0&0&1\\
        0&0&0&0&0&0\\
        0&0&-l&k_{x}&-k_{x}&0\\
        -k_{x}l&k_{x}l&k_{Q}&0&0&1
    \end{bmatrix}^{T},
\end{equation}
where $l$ is the arm length from the center mass to the rotors,  $k_{Q}$ is the rotor torque constant, and $k_x = \frac{\sqrt{2}}{2}$ is a factor that aligns the rotor forces with the principal axes for a X-configuration quadrotor.

As shown in Fig. \ref{fig:blockdiagram}, the goal for control allocation is to solve for $\mathbf{u}$, which will be fed through the actuator mixing block to compute the appropriate rotor speed, and rotor tilt angles. We use the pseudo inverse method to solve this problem given its simplicity \cite{johansen2013control}. Therefore, we implement 
\begin{equation}
            \textbf{u} = \textbf{B}^{\dag}\textbf{v}
\end{equation}
where $\textbf{B}^{\dag}$ is the pseudo-inverse of $\mathbf{B}$.

Next, we use the elements of $\mathbf{u}$ given by \eqref{eq:u} to compute desired rotor speed, and tilt angles as follows
\begin{equation}
    \label{eq:totalThrust}
    T_{i} = \sqrt{(f_{lon}^{\mathcal{B}_{i}})^2+(f_{lat}^{\mathcal{B}_{i}})^2+(f_{ver}^{\mathcal{B}_{i}})^2},
\end{equation}
\begin{equation}
    \label{ew:rotorSpeed}
    \Omega_i = \sqrt{\frac{T_i}{k_T}}, 
\end{equation}

\begin{equation}
    \label{eq:tiltAngleRoll}
    \beta_{i} = \arctan{\frac{f_{lat}^{\mathcal{R}_{i}}}{f_{ver}^{\mathcal{R}_{i}}}}, \quad i = 1,2,3,4.
\end{equation}

These actuator inputs are then used to command the motors and servos, where the dynamics of the vehicle will dictate how the vehicle will respond to the given commands. 

\section{EXPERIMENTS}
\begin{figure*}
    \centering
    \includegraphics[width=0.48\linewidth]{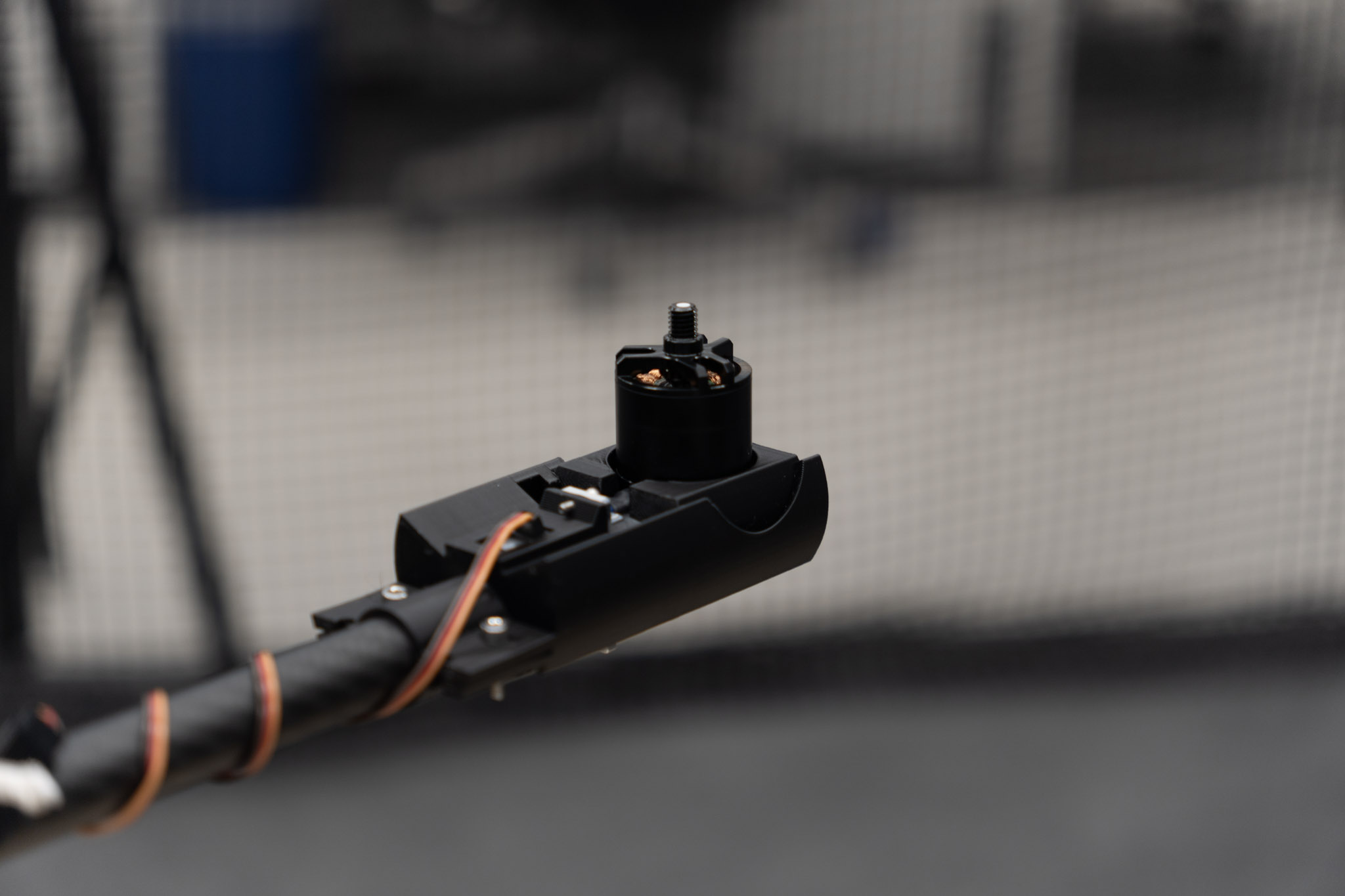}
    \includegraphics[width=0.48\linewidth]{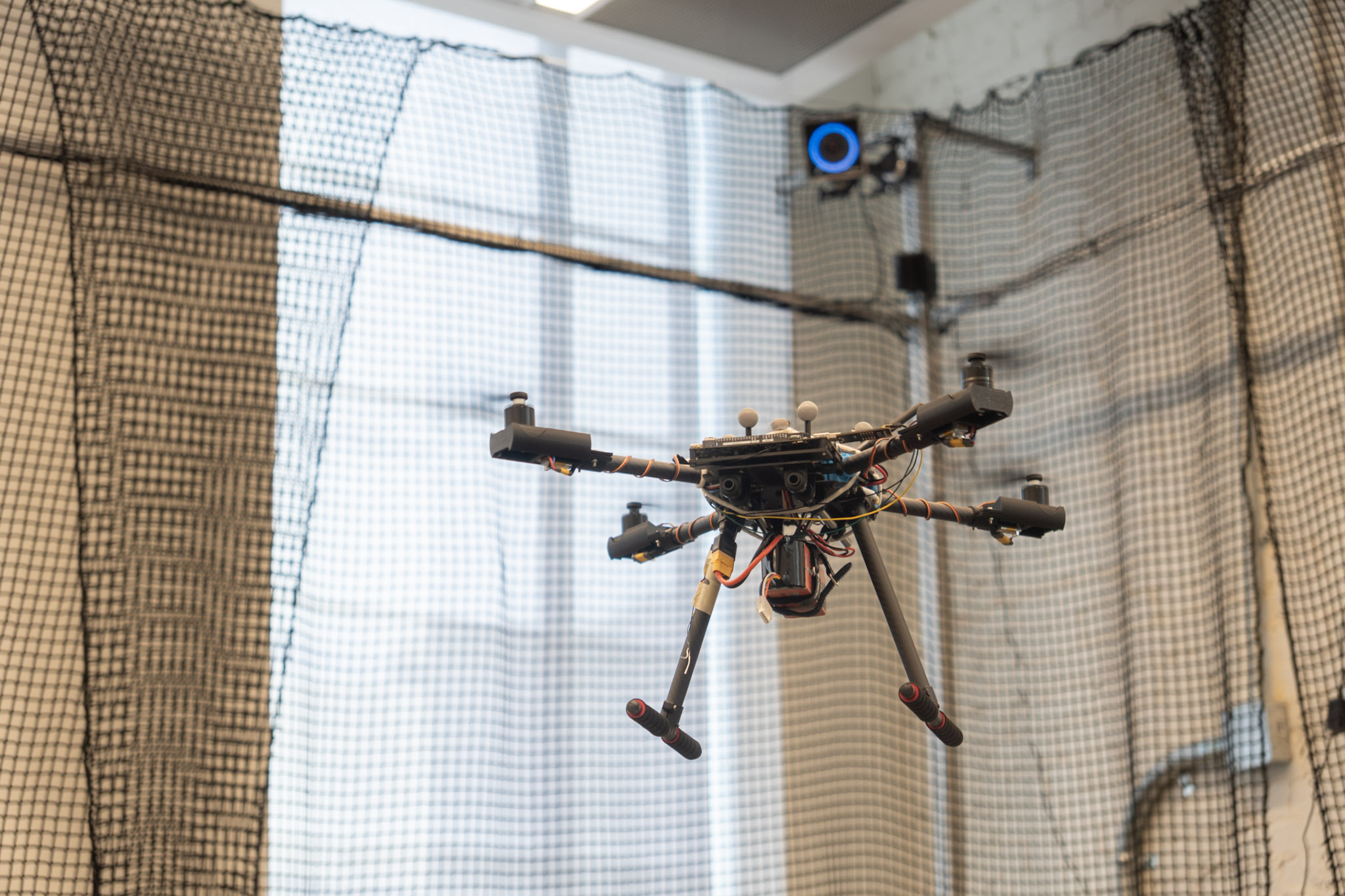}\\
    \caption{Illustrations of the gimballed rotor mechanism and its application in flight. (Left) Close-up view of the gimballed rotor mechanism mounted on the quadrotor (propeller removed for clarity). (Right) The omnidirectional quadrotor in flight, utilizing gimballed rotor mechanisms.}
    \label{fig:experimentalPlatform}
\end{figure*}

\subsection{Prototype Parameters}

We used the proposed gimballed rotor mechanism to modify a Holybro X500 quadrotor and build an omnidirectional quadrotor. Figure~\ref{fig:experimentalPlatform} shows the final fabricated gimbals and the resulting omnidirectional quadrotor.  
Table~\ref{tb:prototypeComponents} presents the bill of materials for the vehicle.  
To control the vehicle, we used a Pixhawk 6C flight controller running the modified PX4 firmware, as described in Section~\ref{se:control}.

\begin{table}[t]
\centering
\caption{Bill of Materials for omnidirectional quadrotor built by the gimballed rotor mechanisms}\label{tb:prototypeComponents}
\begin{tabular}{l c c c} 
\toprule
 Sub-system & Qty & Unit Mass (g) & Total Mass (g) \\ 
\midrule
 Gimballed Rotor    & 4 & 55  & 220 \\ 
 Structural Frame   & 1 & 610 & 610 \\
 Power             & 1 & 500 & 500 \\
 Flight Control    & 1 & 40  & 40 \\
 Flight Computer   & 1 & 175 & 175 \\
 Communications    & 1 & 30  & 30 \\
 Propulsion        & 4 & 75  & 300 \\
 Miscellaneous     & - & -   & 170 \\
\midrule
 Assembled         & - & -   & 2045 \\ 
\bottomrule
\end{tabular}
\end{table}

\subsection{Flight Experiments}
In this section, a level-flight trajectory tracking test is conducted to demonstrate the decoupling of position and attitude. We performed two autonomous flight tests using four waypoints arranged in a square pattern. In the first test, the tilting mechanism was disabled to illustrate how a conventional quadrotor performs in comparison to the second test, where the gimballed rotors were enabled. The tests were conducted indoors, utilizing the OptiTrack motion capture system for position feedback and the built-in inertial measurement unit of the Pixhawk 6C for attitude feedback.

\subsubsection{Attitude Plots}
\begin{figure}[t]
    \centering
        \includegraphics[trim={0cm, 0cm, 0cm, 0cm}, clip, width = \linewidth]{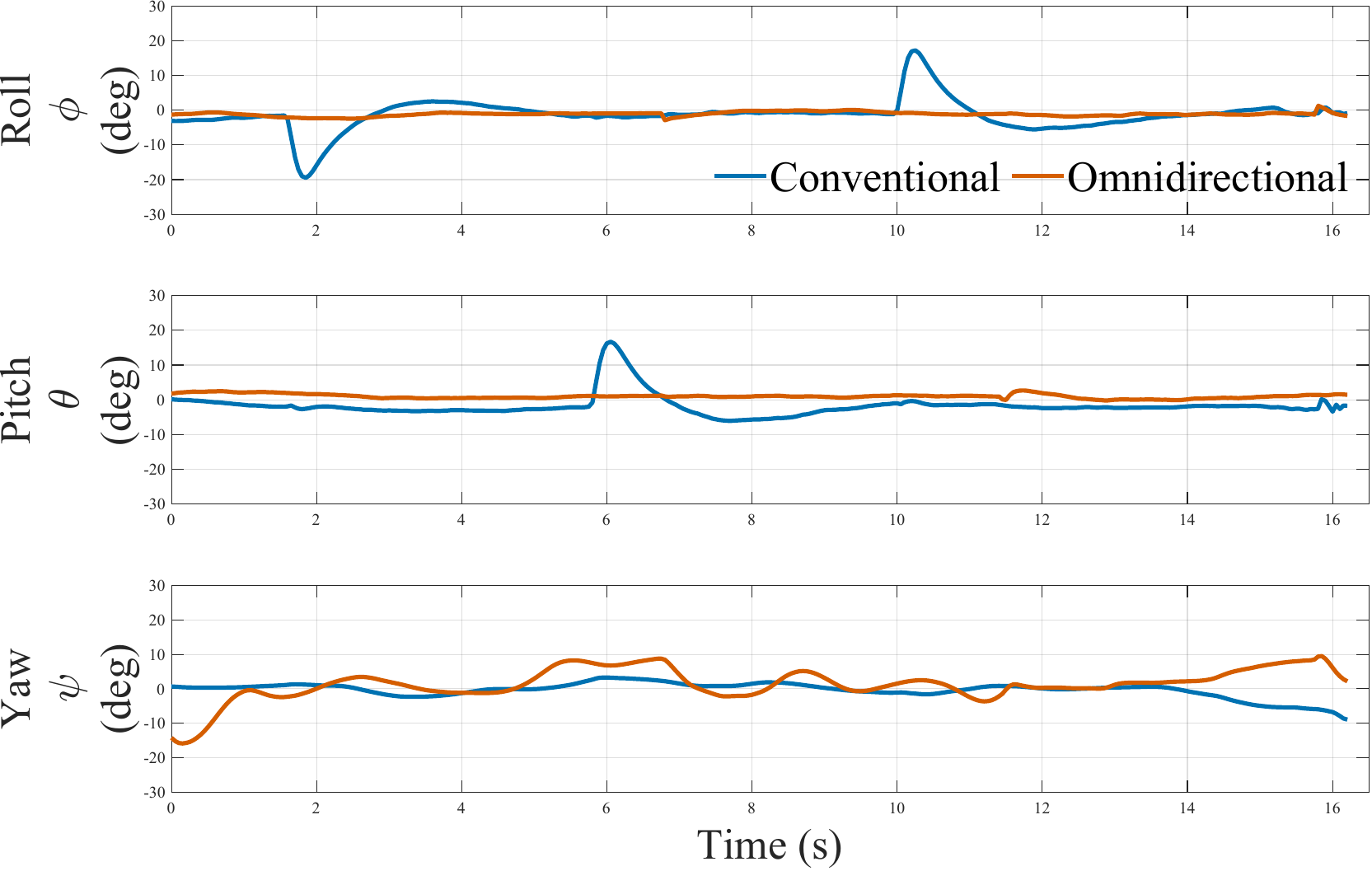}
    \caption{Attitude with respect to time for both the conventional and omnidirectional quadrotor test }
    \label{fig:attitude}
    \end{figure}
  
Figure~\ref{fig:attitude} presents the corresponding roll, pitch, and yaw in the yaw-pitch-roll sequence for both the conventional and omnidirectional quadrotors. It is evident that the gimballed rotors effectively maintain a level attitude throughout the mission. The abrupt increases in roll and pitch occur at waypoints where the vehicle transitions perpendicular to its current path. 

Regarding yaw performance, oscillations can be observed in the omnidirectional quadrotor. This behavior is likely due to aerodynamic interference, which was assumed to be minimal in the model. Additionally, the initially large yaw value can be attributed to a bias introduced when the vehicle was positioned before takeoff.

\subsubsection{Actuators}
    \begin{figure}[t]
    \centering
        \includegraphics[trim={0cm, 0cm, 0cm, 0cm}, clip, width = \linewidth]{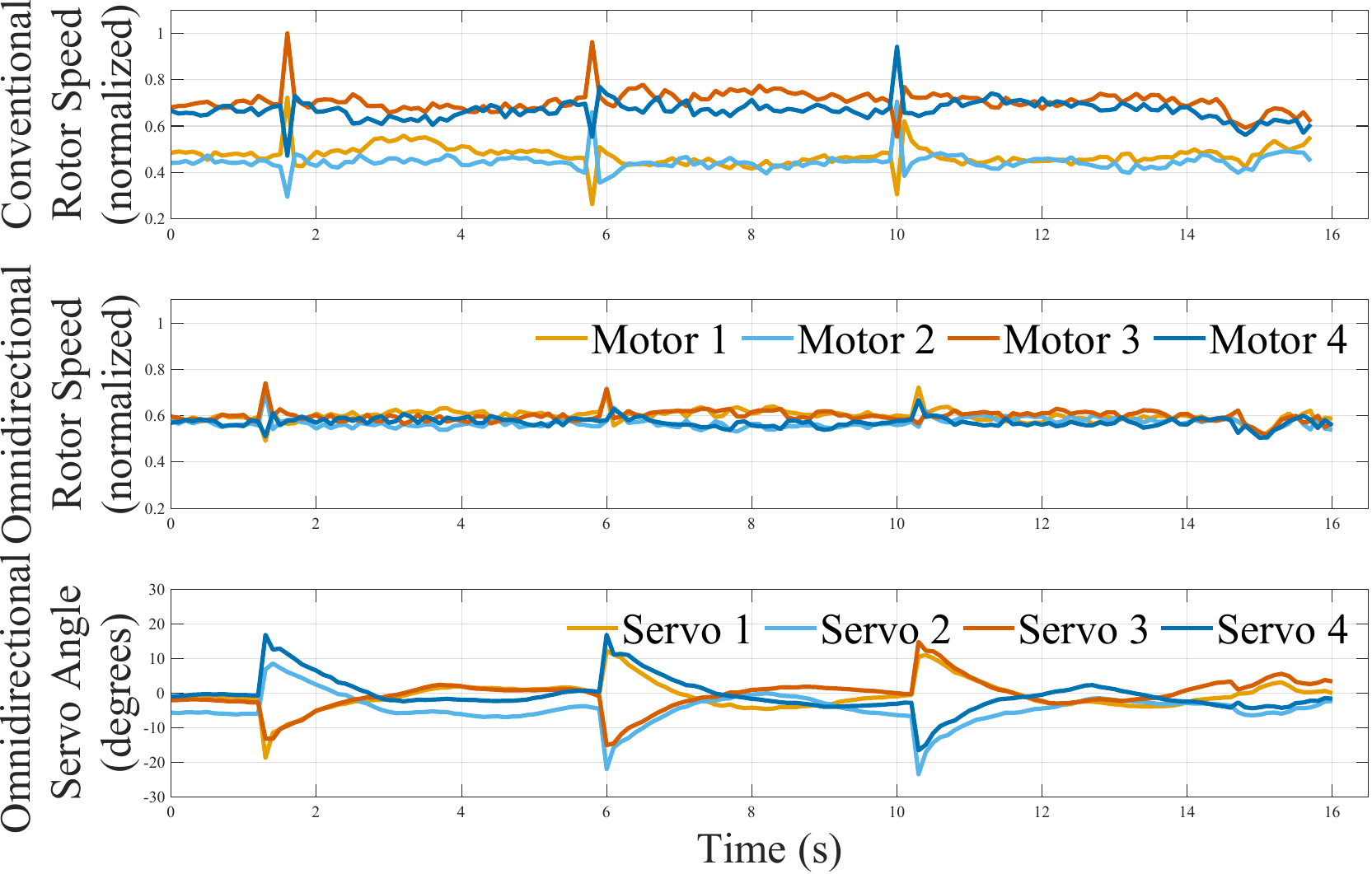}
    \caption{Actuator Inputs with respect to time for both the conventional and omnidirectional quadrotor test}
    \label{fig:actuators}
    \end{figure}

Figure~\ref{fig:actuators} presents the normalized motor speed and tilting angles throughout the flight test. For the conventional quadrotor, it can be observed that the counterclockwise propellers operate at a significantly higher average speed of 0.7 compared to the clockwise propellers, which rotate at approximately 0.45. This speed difference would typically induce a yaw motion; however, as shown in Fig.~\ref{fig:attitude}c, the yaw remains close to zero. This indicates that the vehicle exhibits a natural yaw tendency due to the uneven weight distribution caused by hardware placement.

For the omnidirectional quadrotor, a sinusoidal pattern in the motor speeds is observed, contributing to the yaw oscillation. The tilting mechanism functions as expected, with different pairs of rotors working in coordination to generate the required forces in the $x$- and $y$-directions, which define the primary path of the desired trajectory.

\section{CONCLUSION}

In this paper, we presented the design of a gimballed rotor mechanism as a modular and efficient approach to constructing omnidirectional quadrotors. We also derived the mathematical model of the vehicle’s flight dynamics and detailed the necessary modifications to the PX4 firmware to ensure reliable operation. The fabrication of our prototype and real-world flight demonstrations validated the effectiveness of the proposed design, demonstrating its feasibility as a practical approach for achieving omnidirectional flight without extensive structural modifications. Compared to existing designs, the proposed solution is modular, lightweight, cost-effective, and easy to repair, making it feasible to construct using readily available components and standard fabrication tools. Future work may focus on extension to dual-axis gimballed rotors and implementing a dedicated controller instead of the cascaded control architecture of PX4.

\bibliographystyle{IEEEtran}
\bibliography{References}
\balance
\end{document}